\documentclass[english]{article}

\usepackage{hyperref}
\usepackage{url}
\usepackage{graphicx} 
\usepackage{subfigure}
\usepackage{multirow}
\usepackage{amsthm}
\usepackage{amsmath,amssymb,bm}
\usepackage[ruled] {algorithm2e}
\usepackage{graphicx}
\usepackage{wrapfig}
\newtheorem{definition}{Definition}

\newtheorem{theorem}{Theorem}
\newtheorem{lemma}{Lemma}

\usepackage{wrapfig}
\usepackage{lipsum}

\author{
  {\sc Yingfei Wang}
\thanks{Department of Computer Science,
       Princeton University, 
{\tt yingfei}@{\tt cs.princeton.edu }}
\and
{\sc Chu Wang}
\thanks{Program in Applied and Computational Mathematics,
       Princeton University, 
{\tt   chuw}@{\tt math.princeton.edu }}
\and
{\sc Warren Powell}
\thanks{Department of Operations Research and Financial Engineering,
       Princeton University, 
{\tt   powell}@{\tt princeton.edu }}
}

\title{The Knowledge Gradient with Logistic Belief Models for Binary Classification}

\begin{document}
\date{}

\maketitle

\begin{abstract}
 We consider sequential decision making problems for binary classification scenario in which the learner takes an active role in repeatedly selecting samples from the action pool and receives the binary label of the selected alternatives.  Our problem is motivated by applications where observations are time consuming and/or expensive, resulting in small samples. The goal is to identify the best alternative with the highest response.  We use Bayesian logistic regression to predict the response of each alternative. By formulating the problem as a Markov decision process,  we develop a knowledge-gradient type policy to guide the experiment by maximizing the expected value of information of labeling each alternative and provide a finite-time analysis on the estimated error. Experiments on benchmark UCI datasets demonstrate the effectiveness of the proposed method.
\end{abstract}

\section{Introduction}
There are many classification problems where observations are time consuming and/or expensive.  One example arises in health care analytics, where physicians have to make medical decisions (e.g. a course of drugs, surgery, and expensive tests). Assume that a doctor faces a discrete set of medical choices, and that we can characterize an outcome as a success (patient does not need to return for more treatment) or a failure (patient does need followup care such as repeated operations).  We encounter two challenges.  First, there are very few patients with the same characteristics, creating few opportunities to test a treatment.  Second, testing a medical decision may require several weeks to determine the outcome.  This creates a situation where experiments (evaluating a treatment decision) are time consuming and expensive, requiring that we learn from our decisions as quickly as possible.

The challenge of deciding which medical decisions to evaluate can be modeled mathematically as a sequential decision making problem with binary outcomes. In this setting, we have a budget of measurements that we allocate sequentially to medical decisions so that when we finish our study, we have collected information to maximize our ability to identify the best medical decision with the highest response (probability of success). Scientists can draw on an extensive body of literature on the classic design of experiments \cite{morris1970optimal, wetherill1986sequential, montgomery2008design} whose goal is to decide what observations to make when fitting a function. Yet in the laboratory settings considered in this paper, the decisions need to be guided by a well-defined utility function (that is,  identify the best alternative with the highest probability of success).    This problem also relates to active learning \cite{schein2007active,tong2002support,freund1997selective,settles2010active} in several aspects. In terms of active learning scenarios, our model is most similar to membership query synthesis where  the learner may request labels
for any unlabeled instance in the input space to  learn
a classifier that accurately predicts the labels of new examples. By contrast, our goal is to maximize a utility function such as the success of a treatment. Also, it is typical in active learning not to query a label more than once, whereas we have to live with noisy outcomes, requiring that we sample the same label multiple times. Moreover, the expense of labeling each alternative sharpens the conflicts of learning the prediction and finding the best alternative. Another similar sequential decision making setting is multi-armed bandit problems (e.g. \cite{auer2002finite,bubeck2012regret}). Our work will initially focus on offline settings such as laboratory experiments or medical trials, but the knowledge gradient for offline learning extends easily to online settings \cite{ryzhov2012knowledge}.

There is  a literature studying sequential decision problems to maximize a utility function (e.g., \cite{he2007opportunity,chick2001new,powell2012optimal}). We are particularly interested in a policy that is called the knowledge gradient (KG) that maximizes the expected value of information. After its first appearance for ranking and selection problems \cite{frazier2008knowledge}, KG has been extended to various other belief models (e.g.  \cite{mes2011hierarchical,negoescu2011knowledge,ryzhov2012knowledge,wang2015nested}). Yet there is no KG variant designed for binary classification with parametric models. In this paper, we extend the KG policy to the setting of classification problems under a logistic belief model which introduces the computational challenge of working with nonlinear models.

This paper is organized as follows. We first rigorously establish a sound mathematic model for the problem of sequentially maximizing the response under binary outcome in Section \ref{sec:problem}. We then develop a  recursive Bayesian logistic regression procedure to predict the response of each alternative and  further formulate the problem as a Markov decision process.  In Section 3, we design a knowledge-gradient type policy under a logistic  belief model to guide the experiment  and provide a finite-time analysis on the estimated error. This is different from the PAC (passive) learning bound which relies on the i.i.d. assumption of the examples. Experiments are demonstrated in Section 4.

\section{Problem formulation} \label{sec:problem}
In this section, we state a formal model for our response maximization problem, including transition and objective functions. We then formulate the problem as a Markov decision process.

\subsection{The mathematical model}
We assume that we have a finite set of alternatives $\bm{x}\in \mathcal{X}=\{\bm{x}_1,\dots,\bm{x}_M\}$. The observation of measuring each $\bm{x}$ is a binary outcome $y \in \{-1,+1\}$ with some unknown probability $\text{Pr}[y=+1|\bm{x}]$. Under a limited budget $N$, our goal is to choose the measurement policy $(\bm{x}^1,\dots,\bm{x}^{N})$ and implementation decision $\bm{x}^{N+1}$ that maximizes $\text{Pr}(y=+1|\bm{x}^{N+1})$. We assume a parametric model  where each $\bm{x}$ is a $d$-dimensional vector and  the probability of an example $\bm{x}$ belonging to class $+1$ is given by a nonlinear transformation of an underlying linear function of $\bm{x}$ with a weight vector $\bm{w}$:
$$
\text{Pr}(y=+1|\bm{x},\bm{w})=\sigma(\bm{w}^T\bm{x}),
$$ 
with the sigmoid function $\sigma(a)$ chosen as the logistic function $\sigma(a)=\frac{1}{1+\text{exp}(-a)}.$  

We assume a Bayesian setting  in which we have a multivariate  prior distribution  for the unknown parameter vector $\bm{w}$. At iteration n, we choose an  alternative   $\bm{x}^n$ to measure and observe a binary outcome $y^n$ assuming labels are generated independently given $\bm{w}$. Each alternative can be measured more than once with potentially  different outcomes. Let $\mathcal{D}^n=\{(\bm{x}^i,y^i)\}_{i=1}^n$  denote the previous measured data set  for any $n=1,\dots,N$.  Define the filtration $(\mathcal{F}^n)_{n=0}^N$ by letting $\mathcal{F}^n$ be the sigma-algebra generated by $\bm{x}^1,y^1,\dots, \bm{x}^{n},y^n$. We  use $\mathcal{F}^n$ and $\mathcal{D}^n$ interchangeably.   Measurement and implementation decisions $\bm{x}^{n+1}$ are restricted to be $\mathcal{F}^n$-measurable so that decisions
may only depend on measurements  made in the past. We  use Bayes' rule to form a sequence of posterior predictive distributions $ \text{Pr}(\bm{w}|\mathcal{D}^n)$ for $\bm{w}$ from the prior and the previous measurements. 

The next lemma states the equivalence of using true probabilities and sample estimates when evaluating a policy, where $\Pi$ is the set of policies. The proof is left in the supplementary material.

\begin{lemma}\label{eqv}
Let $\pi \in \Pi$ be a policy, and $\bm{x}^\pi = \arg \max_{\bm{x}} \text{Pr}[y = +1 | \bm{x}, \mathcal{D}^N]$ be the implementation decision after the budget $N$ is exhausted. Then
$$
\mathbb{E}[\text{Pr}(y=+1|\bm{x}^\pi,\bm{w})]=\mathbb{E}[\max_{\bm{x}}\text{Pr}(y=+1|\bm{x},\mathcal{D}^{N})],
$$
where the expectation is taking over the prior distribution of $\bm{w}$.
\end{lemma}

By denoting  $\mathcal{X}^I$ as an implementation policy for selecting an alternative after the measurement budget is exhausted, then $\mathcal{X}^I$ is a mapping from the history $\mathcal{D}^N$ to an alternative $\mathcal{X}^I(\mathcal{D}^N)$. Then as a corollary of Lemma \ref{eqv}, we have \cite{powell2012optimal}
$$\max_{\mathcal{X}^I}\mathbb{E}\big[ \text{Pr}\big(y = +1 | \mathcal{X}( \mathcal{D}^N)\big)\big]= \max_{\bm{x}}\text{Pr}(y = +1 | \bm{x}, \mathcal{D}^N).
$$
In other words, the optimal decision at time $N$ is to go with our final set of beliefs. By the equivalence of using true probabilities and sample estimates when evaluating a policy, while we want to learn the unknown true value $\max_{\bm{x}}\text{Pr}(y=+1|\bm{x})$, we may write our problem's objective  as
\begin{equation}\label{obj}
\max_{\pi \in \Pi} \mathbb{E}^{\pi}[ \max_{\bm{x}}\text{Pr}(y = +1 | \bm{x}, \mathcal{D}^N)].
\end{equation}

\subsection{From logistic regression to Markov decision process formulation}\label{LR}
Logistic regression is  widely used in machine learning for binary classification \cite{hosmer2004applied}. Given a training set $\mathcal{D}=\{(\bm{x}_i,y_i)\}_{i=1}^n$ with $\bm{x}_i$ a $d$-dimensional  vector and $y_i \in \{-1,+1\}$, with the assumption that training labels are generated independently given $\bm{w}$, the likelihood $\text{Pr}(\mathcal{D}|\bm{w})$ is defined as 
$
\text{Pr}(\mathcal{D}|\bm{w}) = \prod_{i=1}^n \sigma(y_i\cdot\bm{w}^T\bm{x}_i).$
In frequentists' interpretation, the weight vector $\bm{w}$ is found by maximizing the likelihood of the training data $\text{Pr}(\mathcal{D}|\bm{w})$.  $l_2$-regularization has been used  to avoid over-fitting with  the estimate of the weight vector $\bm{w}$  given by:
 \begin{equation}\label{RLR}
\min_{\bm{w}} \frac{\lambda}{2}\|\bm{w}\|^2+\sum_{i=1}^n\log(1+\exp(-y_i \bm{w}^T\bm{x}_i)).
\end{equation}
\subsubsection{Bayesian setup}\label{BLR}
Exact Bayesian inference for logistic regression is intractable since the evaluation of the posterior distribution comprises a product of logistic sigmoid functions and the integral in the normalization constant is intractable as well.  With a Gaussian prior on the weight vector, the Laplace approximation can be obtained by finding the mode of the posterior distribution and then fitting a Gaussian distribution centered at that mode (see Chapter 4.5 of \cite{bishop2006pattern}). Specifically, suppose we begin with a Gaussian prior
$
\text{Pr}(\bm{w})= \mathcal{N}(\bm{w}|\bm{m}, \bm{\Sigma}),
$
and we wish to approximate the posterior 
$
\text{Pr}(\bm{w}|\mathcal{D}) \propto \text{Pr}(\mathcal{D}|\bm{w})\text{Pr}(\bm{w}).
$
Define the logarithm of the unnormalized posterior distribution 
\begin{eqnarray}\nonumber \label{LPD}
\Psi(\bm{w}|\bm{m},\bm{\Sigma},\mathcal{D})&=&\log \text{Pr}(\mathcal{D}|\bm{w})+
\log\text{Pr}(\bm{w}) \\
&=& -\frac{1}{2}(\bm{w}-\bm{m})^T\bm{\Sigma}^{-1}(\bm{w}-\bm{m})- \sum_{i=1}^n\log(1+\exp(-y_i \bm{w}^T\bm{x}_i)).
\end{eqnarray}
The Laplace approximation is based on a Taylor expansion to $\Psi$ around its MAP (maximum a posteriori) solution $\hat{\bm{w}}= \arg \max_{\bm{w}}\Psi(\bm{w})$, which defines the mean of the Gaussian. The covariance is then given by the Hessian of the negative log posterior evaluated at $\hat{\bm{w}}$, which takes the form
\begin{equation}\label{LPDD}
(\bm{\Sigma}')^{-1}=-\nabla^2 \Psi(\bm{w})|_{\bm{w}=\hat{\bm{w}}} = \bm{\Sigma}^{-1}+\sum_{i=1}^n p_i(1-p_i)\bm{x}_i\bm{x}_i^T,
\end{equation}
where $p_i = \sigma(\hat{\bm{w}}^T\bm{x}_i)$.
The Laplace approximation results in a normal approximation to the posterior 
\begin{equation}\label{pos}
\text{Pr}(\bm{w}|\mathcal{D}) \approx \mathcal{N}(\hat{\bm{w}},\bm{\Sigma}').
\end{equation}

By substituting an independent normal prior  with $q_i^{-1}$ as the diagonal element of diagonal covariance matrix $\bm{\Sigma}$, the Laplace approximation to the posterior distribution of each weight $w_j$ reduces to
$
\text{Pr}(w_j|\mathcal{D}) \approx \mathcal{N}(\hat{w}_j,q_j^{-1}).
$ Note here if $q_j=\lambda, m_i=0$, the solution of Eq. \eqref{RLR} is the same as the MAP solution of \eqref{LPD}. So an $l_2$-regularized logistic regression can be interpreted as a Bayesian model with a Gaussian prior on the weights with standard deviation $1/\sqrt{\lambda}$.

\subsubsection{Recursive Bayesian logistic update}\label{sec:RBLR}
Our state space is the space of all possible predictive distributions for $\bm{w}$. Starting from a Gaussian prior   $\mathcal{N}(\bm{w}|\bm{m}^0, \bm{\Sigma}^0)$,  after the first $n$ observed data,  the Laplace approximated posterior distribution is $\text{Pr}(\bm{w}|\mathcal{D}^n) \approx \mathcal{N}(\bm{w}|\bm{m}^n, \bm{\Sigma}^n)$ according to \eqref{pos}. We formally define the state space $\mathcal{S}$ to be the cross-product of $\mathbb{R}^d$ and the space of positive semidefinite matrices. At each time $n$, our state of knowledge is thus $S^n=(\bm{m}^n, \bm{\Sigma}^n)$.
Since retraining the logistic model using all the previous data after each new data comes in to update from $S^n$ to $S^{n+1}$ by obtaining the MAP solution of \eqref{LPD} or even with diagonal covariance with  constant diagonal elements  is clumsy, the Bayesian logistic regression can be extended to leverage for recursive model updates after each   of the  training data. 

To be more specific, after the first training data,  the Laplace approximated posterior is $\mathcal{N}(\bm{w}|\bm{m}^1, \bm{\Sigma}^1)$. This serves as a prior on the weights to update the model when the next training data becomes available. In this recursive way of model updating, previously measured data need not  be stored or used for retraining the logistic model.  For the rest of this paper, we focus on independent normal priors (with $ \bm{\Sigma}=\lambda^{-1} \bm{I}$, where $ \bm{I}$ is the identity matrix), which is equivalent to $l_2$-regularized logistic regression, which also offers greater computational efficiency. All the results can be easily generalized to the correlative normal case. By setting the batch size $n=1$ and $\Sigma=\lambda^{-1} \bm{I}$ in Eq. \eqref{LPD} and \eqref{LPDD}, we have the recursive Bayesian logistic regression as in Algorithm \ref{RBLR}.

\begin{algorithm}\label{RBLR}
\caption{Recursive Bayesian Logistic Regression}
\SetKwInOut{Input}{input}\SetKwInOut{Output}{output}

 \Input{Regularization parameter $\lambda > 0$} 
 $m_j=0$,  $q_j=\lambda$. (Each weight $w_j$ has an independent prior $\mathcal{N}(m_j, q_j^{-1})$)\\
 \For{$t=1$ to $T$}{
 ~\\
 Get a new point $(\bm{x}, y)$.\\
Find $\hat{\bm{w}}$ as the maximizer of \eqref{LPD}: $-\frac{1}{2}\sum_{j=1}^d q_i(w_i-m_i)^2 - \log(1+\exp(-y\bm{w}^T\bm{x})).$\\
$m_j=\hat{w}_j$\\
Update $q_i$ according to \eqref{LPDD}:
$q_j \leftarrow q_j +\sigma(\hat{\bm{w}}^T\bm{x})(1-\sigma(\hat{\bm{w}}^T\bm{x})x^2_{j}$.
}
\end{algorithm}

Since $\Psi(\bm{w}|\bm{m},\bm{\Sigma},\mathcal{D})$ is convex in $\bm{w}$, we can tap a wide range of convex optimization algorithms including gradient search, conjugate gradient, an BFGS method (see \cite{wright1999numerical}  for details). Yet when setting $n=1$ and $\Sigma=\lambda^{-1} \bm{I}$ in Eq. \eqref{LPD}, a  stable and efficient algorithm for solving $\arg\max \Psi(\bm{w})=-\frac{1}{2}\sum_{j=1}^d q_i(w_i-m_i)^2 - \log(1+\exp(-y\bm{w}^T\bm{x}))$ can be obtained as follows. First we calculate $$\frac{\partial F}{\partial w_i}=-q_i(w_i-m_i)+\frac{yx_i\exp(-y\mathbf{w}^T\mathbf{x})}{1+\exp(-y\mathbf{w}^T\mathbf{x})}.$$ By setting 
$\partial F/\partial w_i=0$ for all $i$, then by denoting $(1+\exp(y\mathbf{w}^T\mathbf{x}))^{-1}$ as $p$, we have
$$q_i(w_i-m_i)=ypx_i,~~~~i=1,2,\dots,d,$$ and thus
$w_i=m_i+yp\frac{x_i}{q_i}.$
Plugging in these equalities into the definition of $p$, we have
$$\frac{1}{p}=1+\exp\Big{(}y\sum_{i=1}^{d}(m_i+yp\frac{x_i}{q_i})x_i\Big{)}
=1+\exp(y\mathbf{m}^T\mathbf{x})\exp \Big{(}y^2p\sum_{i=1}^d\frac{x_i^2}{q_i}\Big{)}.$$

The left hand side decreases from infinity to 1 and the right hand side increases from 1 when $p$ goes from 0 to 1, therefore the solution exists and is unique in $[0,1]$. By reducing a $d$-dimensional problem to a $1$-dimensional one, the simple bisection method is good enough.

\subsection{Markov decision process formulation}
Our learning problem is a dynamic program that can be formulated as a Markov decision process. By using diagonal covariance matrices, the state space degenerates to  $\mathcal{S} := \mathbb{R}^d \times (0,\infty] ^d$ and  it consists of points $s=(\bm{m},\bm{q})$, where $m_i, q_i$ are the mean and the precision of a normal distribution. We next define the transition function based on the recursive Bayesian logistic regression.
\begin{definition}The transition function $T$: $\mathcal{S}\times\mathcal{X}\times \{-1, 1\}$ is defined as
$$T\Big((\bm{m},\bm{q}), \bm{x},y\Big) = \bigg(\Big(\hat{\bm{w}}(\bm{m}),\bm{q}+ p(1-p)\textbf{diag}(\bm{x}\bm{x}^T)\Big)\bigg),
$$
where $\hat{\bm{w}}(\bm{m})=\arg \min_{\bm{w}}\Psi(\bm{w}|\bm{m},\bm{q})$, $p = \sigma(\hat{\bm{w}}^T\bm{x})$ and $\textbf{diag}(\bm{x}\bm{x}^T)$ is a column vector containing the diagonal elements of $\bm{x}\bm{x}^T$, so that $S^{n+1}=T(S^n, \bm{x},y)$.
\end{definition}

In a dynamic program, the value function is defined as the value of the optimal policy given a particular state
$S^n$ at  time $n$, and may also be determined recursively through Bellman's equation. If the value function can be computed efficiently, the optimal policy may then also be computed from it.  The value function $V^n:\mathcal{S} \mapsto \mathbb{R}$ at time $n=1,\dots,N+1$ is given by   \eqref{obj} as 
$$
V^n(s) := \max_{\pi} \mathbb{E}^{\pi}[ \max_{\bm{x}}\text{Pr}(y = +1 | \bm{x}, \mathcal{F}^N)|S^n=s].
$$

By noting that $\max_{\bm{x}}\text{Pr}(y = +1 | \bm{x},\mathcal{F}^N$) is $\mathcal{F}^N$-measurable and thus the expectation does not depend on policy $\pi$, the terminal value function $V^N$ can be computed directly as 
$$
V^{N+1}(s)=\max_{\bm{x}}\text{Pr}(y = +1 | \bm{x},s), \forall s \in \mathcal{S}.
$$

The value function at times $n=1,\dots,N$, $V^n$, is given recursively by
$$
V^n(s)=\max_{\bm{x}}\mathbb{E}[V^{n+1}(T(s,\bm{x},y))], s \in \mathcal{S}.
$$
\section{Knowledge gradient policy for logistic belief model}
Since the ``curse of dimensionality''
makes direct computation of the value function  intractable, computationally efficient approximate
policies need to be considered.  A computationally  attractive policy for ranking and selection problems is the knowledge
gradient (KG), which is a stationary policy that at the $n$th iteration chooses its ($n+1$)th measurement to maximize the single-period expected increase in value \cite{frazier2008knowledge}.  It enjoys some nice properties, including myopic and asymptotic optimality. After its first appearance, KG has been extended to various belief models (e.g.  \cite{mes2011hierarchical,negoescu2011knowledge,ryzhov2012knowledge,wang2015nested}) for offline learning, and an immediate extension to online learning problems \cite{ryzhov2012knowledge}. Yet there is no KG variant designed for binary classification with parametric models,  primarily because of the complexity of dealing with nonlinear belief models. In what follows, we extend the KG policy to the setting of classification problems under a logistic belief model.
\begin{definition}The knowledge gradient of measuring an alternative $\bm{x}$ while in state $s$ is 
\begin{equation} \label{KG}
\nu_{\bm{x}}^{\text{KG}}(s) := \mathbb{E}[V^{N+1}(T(s,\bm{x},y))-V^{N+1}(s)].
\end{equation}
\end{definition}

Since the label for alternative $\bm{x}$ is not known at the time of selection,  the expectation is computed conditional on the current model specified by $s=(\bm{m},\bm{q})$. Specifically,
given a state $s=(\bm{m},\bm{q})$, the label $y$ for an alternative $\bm{x}$ follows from a Bernoulli distribution with a predictive distribution 
\begin{eqnarray}\label{predictD}
\text{Pr}(y=+1|\bm{x},s) =\int \text{Pr}(y=+1|\bm{x},\bm{w})\text{Pr}(\bm{w}|s)\text{d}\bm{w}
= \int \sigma(\bm{w}^T\bm{x})p(\bm{w}|s)\text{d}\bm{w}.
\end{eqnarray}

We have
\begin{eqnarray*}
\mathbb{E}[V^{N+1}(T(s,\bm{x},y))] &=& \text{Pr}(y=+1|\bm{x},s)V^N(T(s, \bm{x},+1))+ \text{Pr}(y=-1|\bm{x},s)V^N(T(s, \bm{x},-1))\\
&=&\text{Pr}(y=+1|\bm{x},s)\cdot  \max_{\bm{x}'}\text{Pr}(y = +1 | \bm{x}',T(s,\bm{x},+1))\\
&&+\text{Pr}(y=-1|\bm{x},s)\cdot  \max_{\bm{x}'}\text{Pr}(y = +1 | \bm{x}',T(s,\bm{x},-1)).
\end{eqnarray*}

The knowledge gradient policy suggests at each time $n$ selecting the alternative that maximizes $\nu_{\bm{x}}^{\text{KG}}(s^{n-1})$ where  ties are broken randomly. The same  optimization procedure as in recursive Bayesian logistic regression needs to be conducted for calculating the transition functions $T(s,\bm{x},\cdot)$.

The predictive distribution $ \int \sigma(\bm{w}^T\bm{x})p(\bm{w}|s)\text{d}\bm{w}$ cannot be evaluated exactly using the logistic function in the role of the sigmoid $\sigma$.  An approximation procedure is deployed as follows.
Denoting $a=\bm{w}^T\bm{x}$ and $\delta(\cdot)$ as the Dirac delta function, we have $\sigma(\bm{w}^T\bm{x})=\int \delta(a-\bm{w}^T\bm{x})\sigma(a)\text{d}a.$
Hence 
$$\int \sigma(\bm{w}^T\bm{x})p(\bm{w}|s) \text{d}\bm{w}=\int \sigma(a)p(a)\text{d}a,$$
where 
$p(a)=\int \delta(a-\bm{w}^T\bm{x})p(\bm{w}|s)  \text{d}\bm{w}.$
Since $p(\bm{w}|s) = \mathcal{N}(\bm{m},\bm{q}^{-1})$ is Gaussian, the marginal distribution $p(a)$ is also Gaussian. We can evaluate $p(a)$ by calculating the mean and covariance of this distribution \cite{bishop2006pattern}. We have
\begin{eqnarray*}
\mu_a&=&\mathbb{E}[a]=\int p(a)a \text{ d}a = \int p(\bm{w}|s)\bm{w}^T\bm{x} \text{ d}\bm{w}=\hat{\bm{w}}^T\bm{x},\\
\sigma_a^2&=& \int p(\bm{w}|s) \big((\bm{w}^T\bm{x})^2-(\bm{m}^T\bm{x})^2 \big) \text{ d}\bm{w}=\sum_{j=1}^d q_j^{-1} x_j^2.
\end{eqnarray*} Thus $\int \sigma(\bm{w}^T\bm{x})p(\bm{w}|s) \text{d}\bm{w}=\int \sigma(a)p(a)\text{d}a=\int \sigma(a) \mathcal{N}(a|\mu_a, \sigma^2_a) \text{d}a.$

For a logistic function, in order to obtain the best approximation \cite{barber1998ensemble,spiegelhalter1990sequential}, we approximate $\sigma(a)$ by $\Phi(\alpha a)$ with $\alpha=\pi/8$. Denoting $\kappa(\sigma^2)=(1+\pi \sigma^2/8)^{-1/2}$ ,  we have
$$
\text{Pr}(y=+1|\bm{x},s)=\int \sigma(\bm{w}^T\bm{x})p(\bm{w}|s) \text{d}\bm{w} \approx \sigma(\kappa(\sigma^2)\mu).
$$

We summarize the decision rule of the knowledge gradient policy  at each iteration in Algorithm \ref{KG}.

\begin{algorithm}\label{KG}
\caption{Knowledge Gradient Policy for Logistic Belief Model}
\SetKwInOut{Input}{input}\SetKwInOut{Output}{output}

 \Input{$m_j$, $q_j$  (Each weight $w_j$ has an independent prior $\mathcal{N}(m_j, q_j^{-1})$)}
 
 \For{$\bm{x}$ in $\mathcal{X}$}{
 ~\\
Let $\Psi(\bm{w},y)=-\frac{1}{2}\sum_{j=1}^d q_i(w_i-m_i)^2 - \log(1+\exp(-y\bm{w}^T\bm{x}))$\\
$\hat{\bm{w}}_{+}=\arg \max_{\bm{w}}\Psi(\bm{w},+1)$ \\
$\hat{\bm{w}}_{-}=\arg \max_{\bm{w}}\Psi(\bm{w},-1)$ \\
$\mu=\bm{m}^T\bm{x}$\\
$\sigma^2=\sum_{j=1}^d q_j^{-1} x_j^2$\\
Define $\mu_{+}(\bm{x}')=\hat{\bm{w}}_{+}^T\bm{x}'$,  $\mu_{-}(\bm{x}')=\hat{\bm{w}}_{-}^T\bm{x}'$\\
Define $\sigma^2_{\pm}(\bm{x}')= \sum_{j=1}^d\Big( q_j+\sigma(\hat{\bm{w}}^T_{\pm}\bm{x}')(1-\sigma(\hat{\bm{w}}_{\pm}^T\bm{x}')x^2_{j} \Big) (x'_j)^2$\\
$\tilde{\nu}_{\bm{x}}=\sigma(\kappa(\sigma^2)\mu)\cdot \max_{\bm{x'}}\sigma \big(\kappa(\sigma_+^2(\bm{x}'))\mu_{+}(\bm{x'})\big)+\big(1-\sigma(\kappa(\sigma^2)\mu)\big)\cdot \max_{\bm{x'}}\sigma \big(\kappa(\sigma_-^2(\bm{x}'))\mu_{-}(\bm{x'})\big)$\\
}
$\bm{x}^{\text{KG}} = \arg \max_{\bm{x}}\tilde{\nu}_{x}$
\end{algorithm}

We close this section by presenting the following finite-time bound on the MSE
of the estimated weight with the proof in the supplement. Without loss of generality, we assume $\|\bm{x}\|_2 \le 1$, $\forall{\bm{x}\in \mathcal{X}}$.
\begin{theorem}

Let $\mathcal{D}^n$ be the $n$ measurements produced by the KG policy and $\bm{w}^n=\arg\max_{\bm{w}}\Psi(\bm{w}|\bm{m}^0, \Sigma^0)$ with the prior distribution  $\text{Pr}(\bm{w}^*)= \mathcal{N}(\bm{w}^*|\bm{m}^0, \bm{\Sigma}^0).$ Then with probability $P_d(M)$, the expected error of $\bm{w}^n$ is bounded as
$$\mathbb{E}_{ \bm{y}\sim \mathcal{B}(\mathcal{D}^n,\bm{w}^*)}||\bm{w}^n-\bm{w}^*||_2\le\frac{C_{min}+\lambda_{min}\big{(}\bm{\Sigma}^{-1}\big{)}}{2},$$
where the distribution  $\mathcal{B}(\mathcal{D}^n,\bm{w}^*)$ is the vector Bernoulli distribution
$Pr(y^i=+1)=\sigma(\bm{w}^{*T}\bm{x}^i)$,
$P_d(M)$ is the probability of a d-dimension standard normal random variable appears in the ball with radius $M =\frac{1}{8}\frac{\lambda_{min}^2}{\sqrt{\lambda_{max}}}$ and
$C_{min}= \lambda_{min}
\Big{(}\frac{1}{n}\sum_{i=1} \sigma(\bm{w}^{*T}\bm{x}^i)\big{(}1-\sigma(\bm{w}^{*T}\bm{x}^i)\big{)}\bm{x}^i(\bm{x}^i)^T \Big{)}.$
\end{theorem}
In the special case where $\bm{\Sigma}^0=\lambda^{-1}\bm{I}$, we have $\lambda_{max}=\lambda_{min}=\lambda$ and  $M=\frac{\lambda^{3/2}}{8}$.
The bound holds with higher probability  $P_d(M)$ with Iarger $\lambda$. This is natural since a larger $\lambda$ represents a normal distribution with narrower bandwidth, resulting in a more concentrated $\bm{w}^*$ around $\bm{m}^0$.

\section{Experimental results}
We evaluate the proposed method on both synthetic datasets and the UCI machine learning repository  \cite{Lichman:2013} which includes classification problems drawn from settings including fertility, glass identification, blood transfusion, survival, breast cancer  (wpbc), planning relax and climate model  failure. We first  analyze the behavior of the KG policy and then compare it to state-of-the-art active learning algorithms. On synthetic datasets, we randomly generate a set of $M$ $d$-dimensional alternatives  $\bm{x}$ from [-3,3]. We conduct  experiments in a Bayesian fashion where at each run we sample a true $d+1$-dimensional weight vector $\bm{w}^*$ from the prior distribution $w^*_i \sim \mathcal{N}(0, \lambda)$. The $+1$ label for each alternative $\bm{x}$ is generated with probability $\sigma(w^*_0+\sum_{j=1}^d w^*_dx_d)$. For each  UCI dataset, we use all the data points as the set of alternatives with their original attributes. We then simulate their labels using a weight vector $\bm{w}^*$. This weight vector could have been chosen arbitrarily, but it was in fact a perturbed version of the weight vector trained through logistic regression on the original dataset. All the policies start with the same one randomly selected example per class.

\subsection{Behavior of the KG policy}
\begin{figure}[htp]
    \centering
    \hspace*{-0.4cm}
    \begin{tabular}{cccc}
            \includegraphics[width=0.25\textwidth]{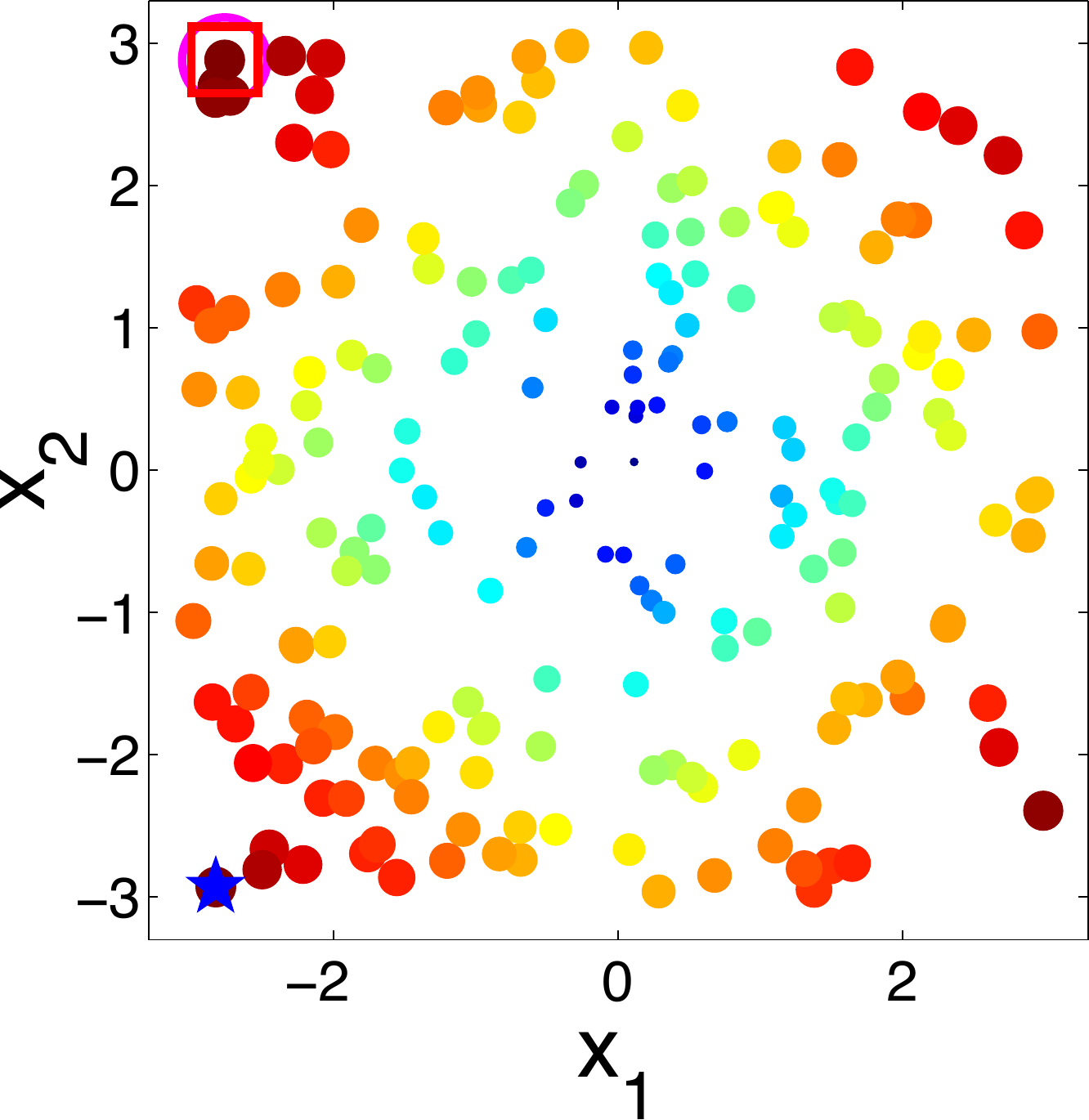}
  \includegraphics[width=0.25\textwidth]{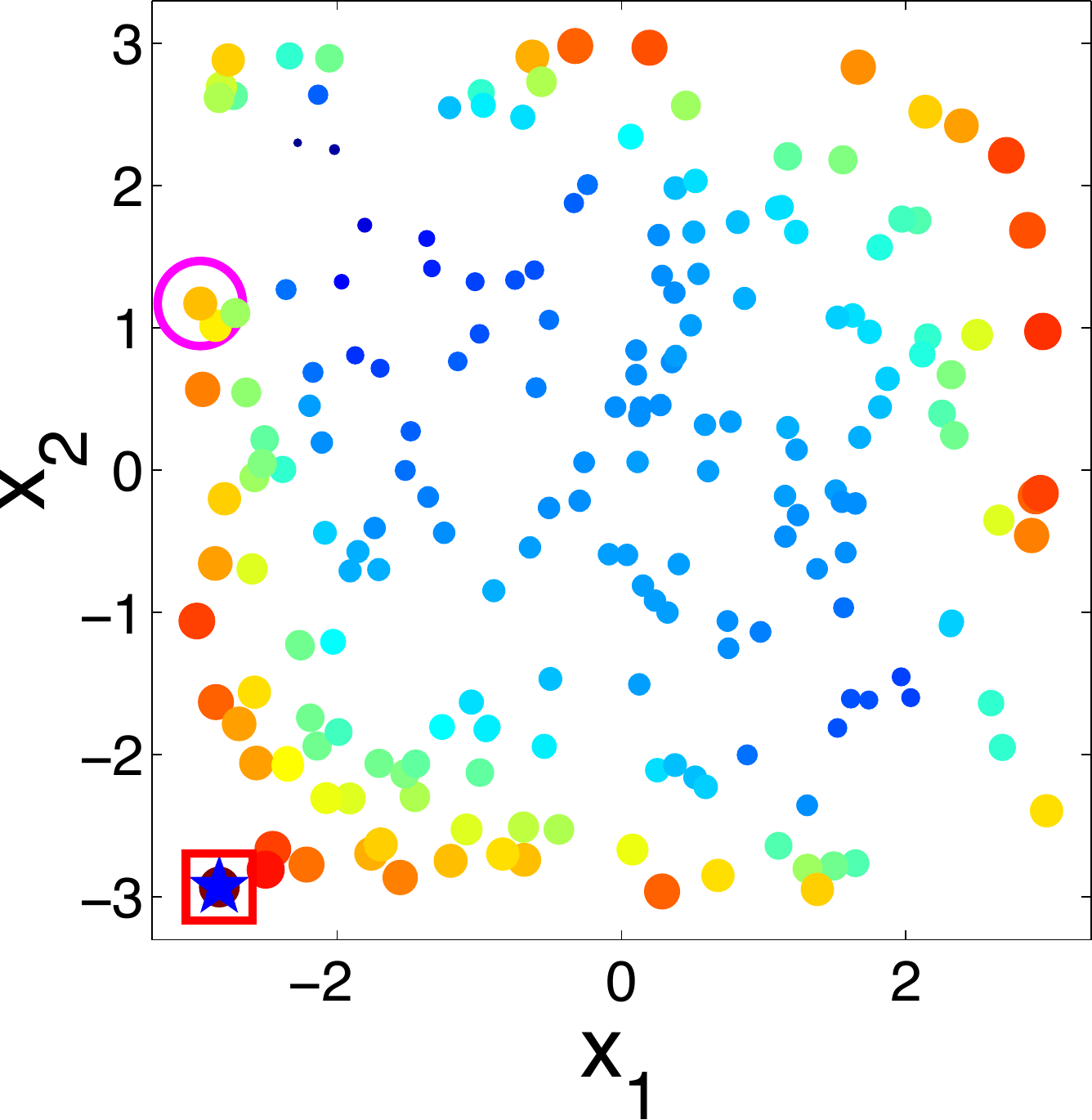} 
 \includegraphics[width=0.25\textwidth]{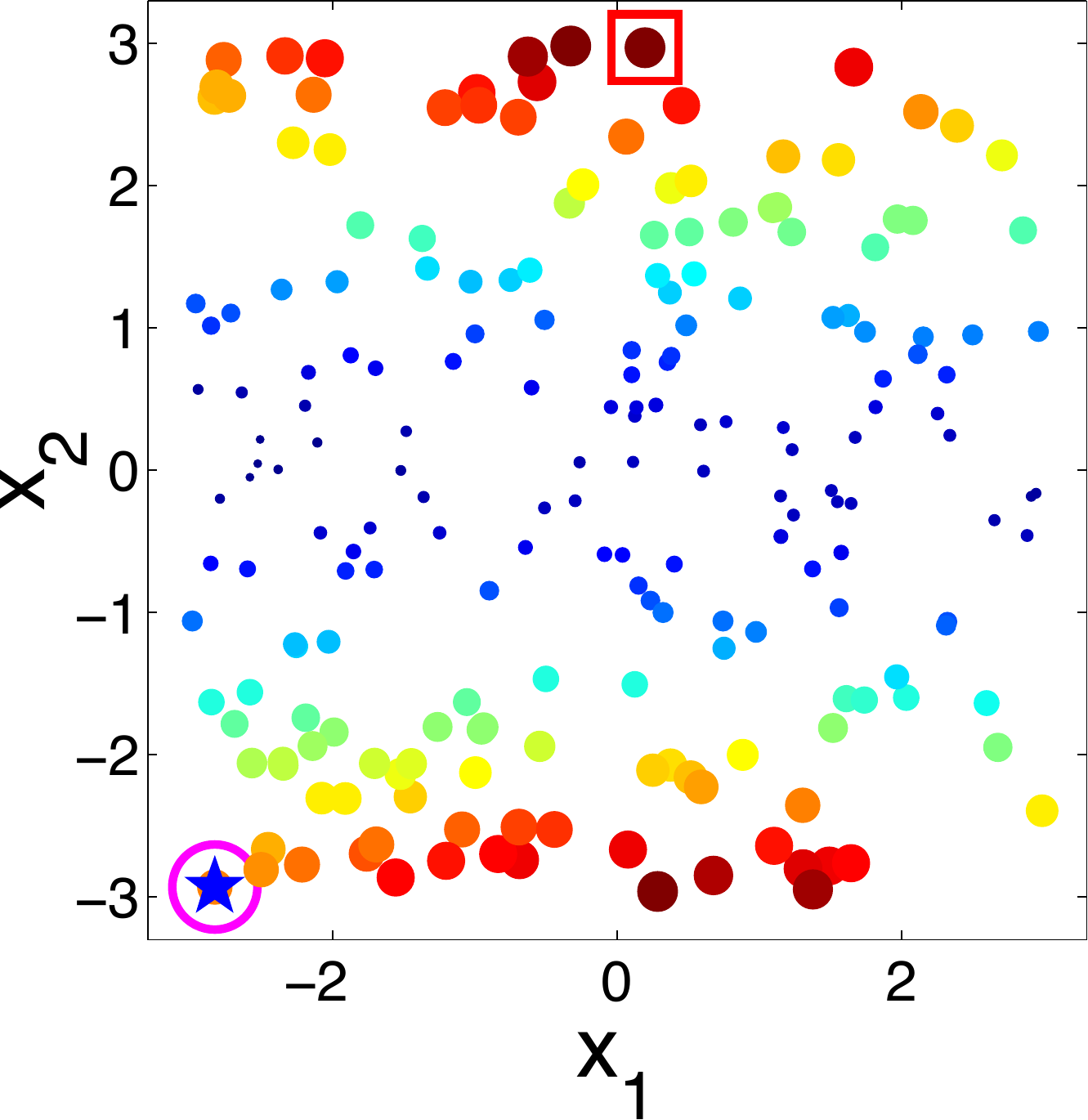} 
      \includegraphics[width=0.25\textwidth]{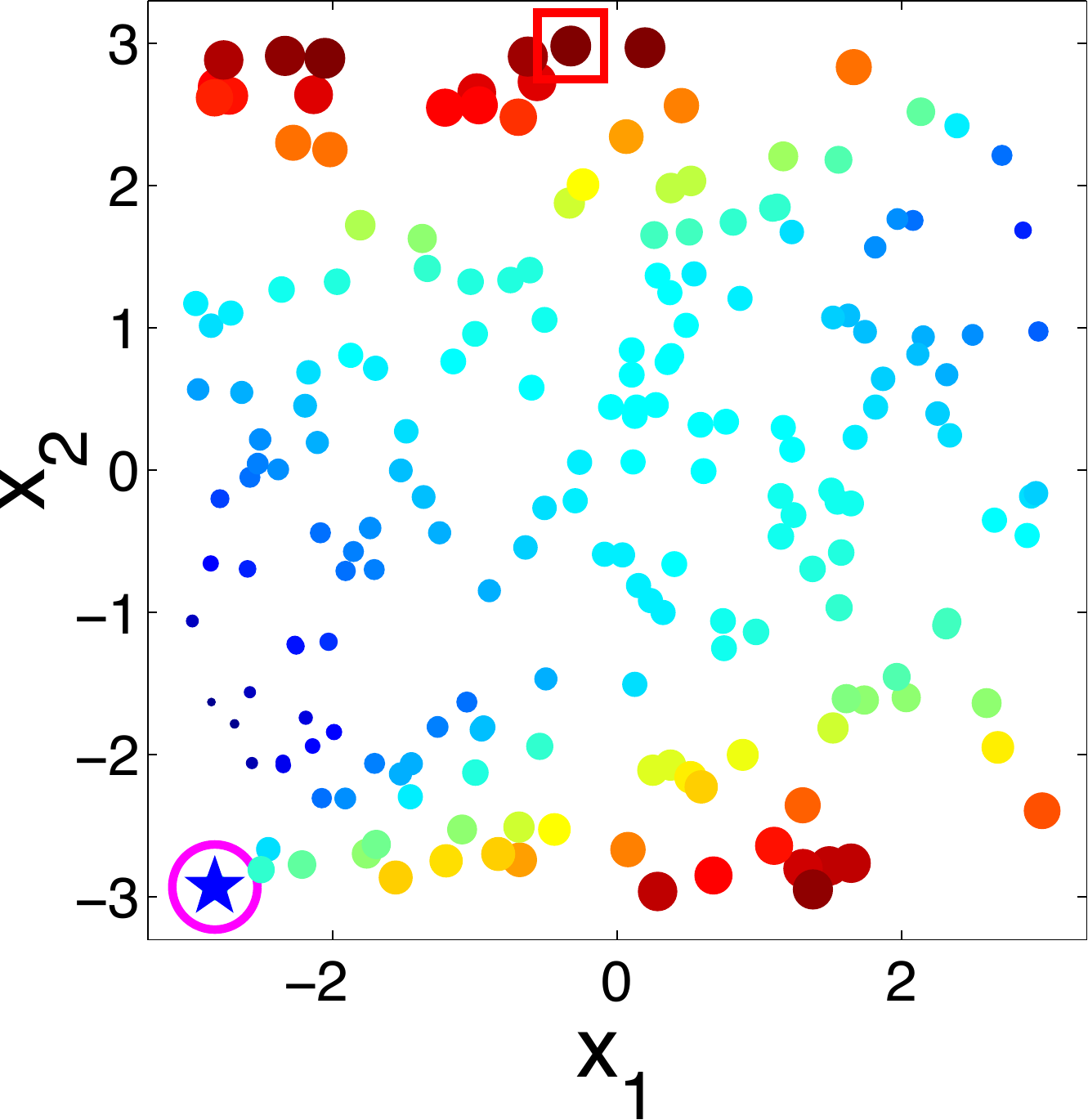}

    \end{tabular}
    \caption{The scatter plots illustrate the KG values at 1-4 iterations from left to right with both the color and the size  reflecting the magnitude. The star, the red square and pink circle indicate the true best alternative, the alternative to be selected and  the implementation decision, respectively. \label{2d}}
\end{figure}

To better understand the behavior of the KG policy,  Fig. \ref{2d} shows the snapshot of the KG policy at each iteration on a $2$-dimensional  synthetic dataset ($M=200$) in one run. The scatter plots show the KG values with both the color and the size of the point reflecting the KG value of the corresponding alternative. The star denotes the true alternative with the largest response. The red square is the alternative with the largest KG value. The pink circle is the implementation decision that maximizes the response under current estimation of $\bm{w}^*$ if the budget is exhausted after that iteration.

It can be seen from the figure that the KG policy finds the true best alternative after only three measurements, reaching out to different alternatives to improve its estimates.  We can infer from Fig. \ref{2d} that the KG policy tends to choose alternatives near the boundary of the region.
This criterion is natural since in order to find the true maximum, we need to get enough information about $\bm{w}^*$ and 
estimate well the probability of points near the true maximum which appears near the boundary. On the other hand,  in a logistic model with labeling noise, a data $\bm{x}$ with small $\bm{x}^T\bm{x}$ inherently brings  little  information as pointed out in \cite{zhang2000value}. For an extreme example, when $\bm{x}=\bm{0}$ the label is always completely random for any $\bm{w}$ since $\text{Pr}(y=+1|\bm{w},\bm{0}) \equiv 0.5$.  This is an issue when perfect classification is not achievable. So it is essential to label a data with larger $\bm{x}^T\bm{x}$ that has the most potential to  enhance its confidence non-randomly. 
\begin{wrapfigure}{r}{0.6\textwidth}
  \begin{center}
    \includegraphics[width=0.6\textwidth]{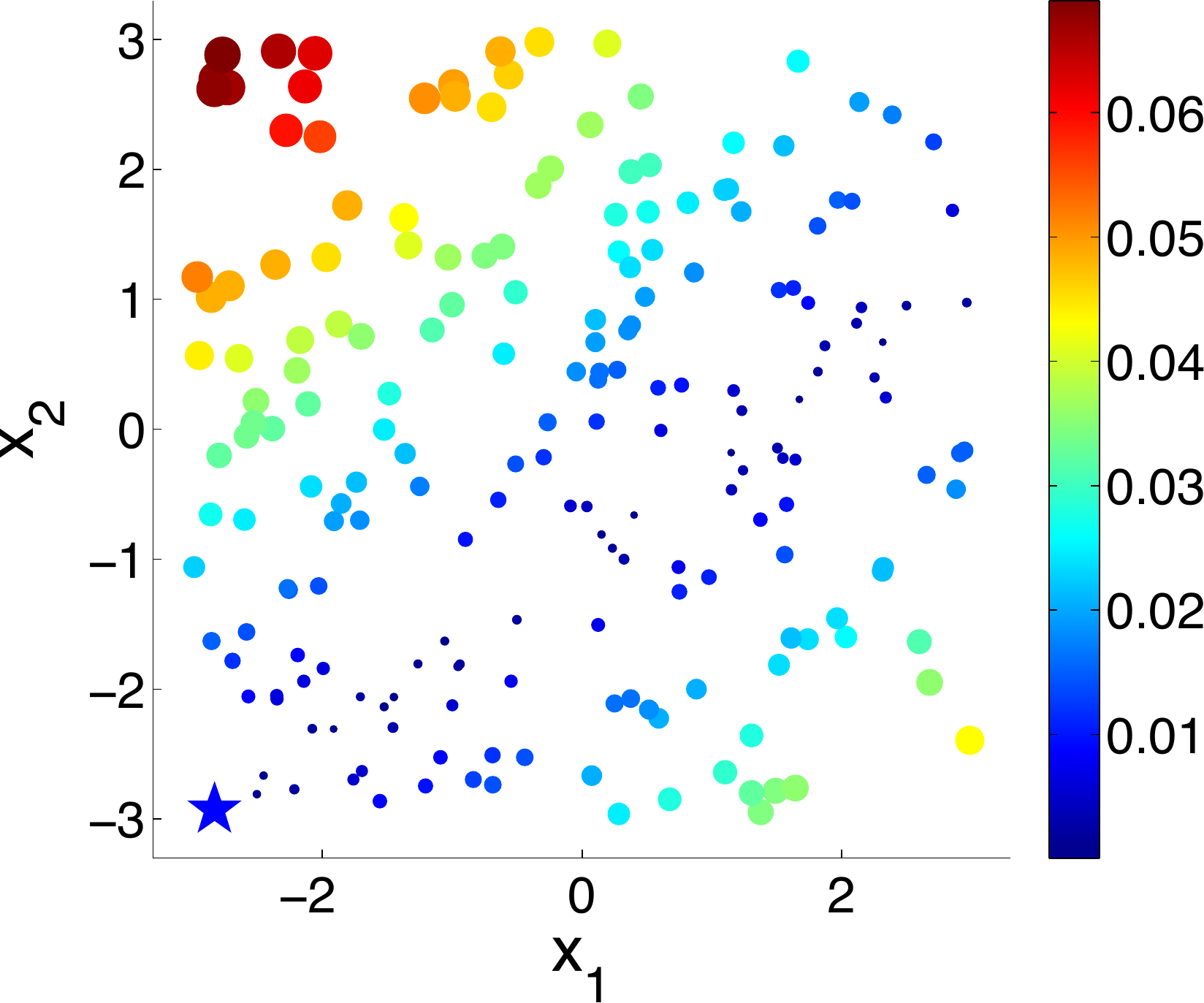}
  \end{center}
\caption{Absolute  error. \label{Abs}}
\end{wrapfigure}

Also depicted in Fig. \ref{Abs} is the absolute class distribution error of each alternative, which is the absolute difference between the predictive probability of class $+1$ under current estimate and the true probability after $6$ iterations. We see that the probability  at the true maximum is well approximated, while moderate error in
the estimate is located away from this region of interest. We also provide the analysis on a  3-dimensional dataset in the supplement.

\subsection{Comparison with other policies} 
Recall that our goal is to maximize the expected response of the implementation decision. We define the Opportunity Cost (OC) metric as the expected response of the implementation decision $\bm{x}^{N+1}:=\arg\max_{\bm{x}} \text{Pr}(y=+1|\bm{x},\bm{w}^N)$ compared to the true maximal response under weight $\bm{w}^*$: 
$$\text{OC}:=\max_{\bm{x}\in \mathcal{X}}\text{Pr}(y=+1|\bm{x},\bm{w}^*)-\text{Pr}(y=+1|\bm{x}^{N+1},\bm{w}^*).$$
Note that the opportunity cost is always non-negative and the smaller the better.  
To make a fair comparison, on each run, all the time-$N$ labels of all  the alternatives are randomly pre-generated according to the weight vector $\bm{w}^*$ and shared across all the competing policies.  

\begin{figure}[htp!]
   \centering
    \begin{tabular}{ccc}
\subfigure[fertility]{
 \includegraphics[width=0.29\textwidth]{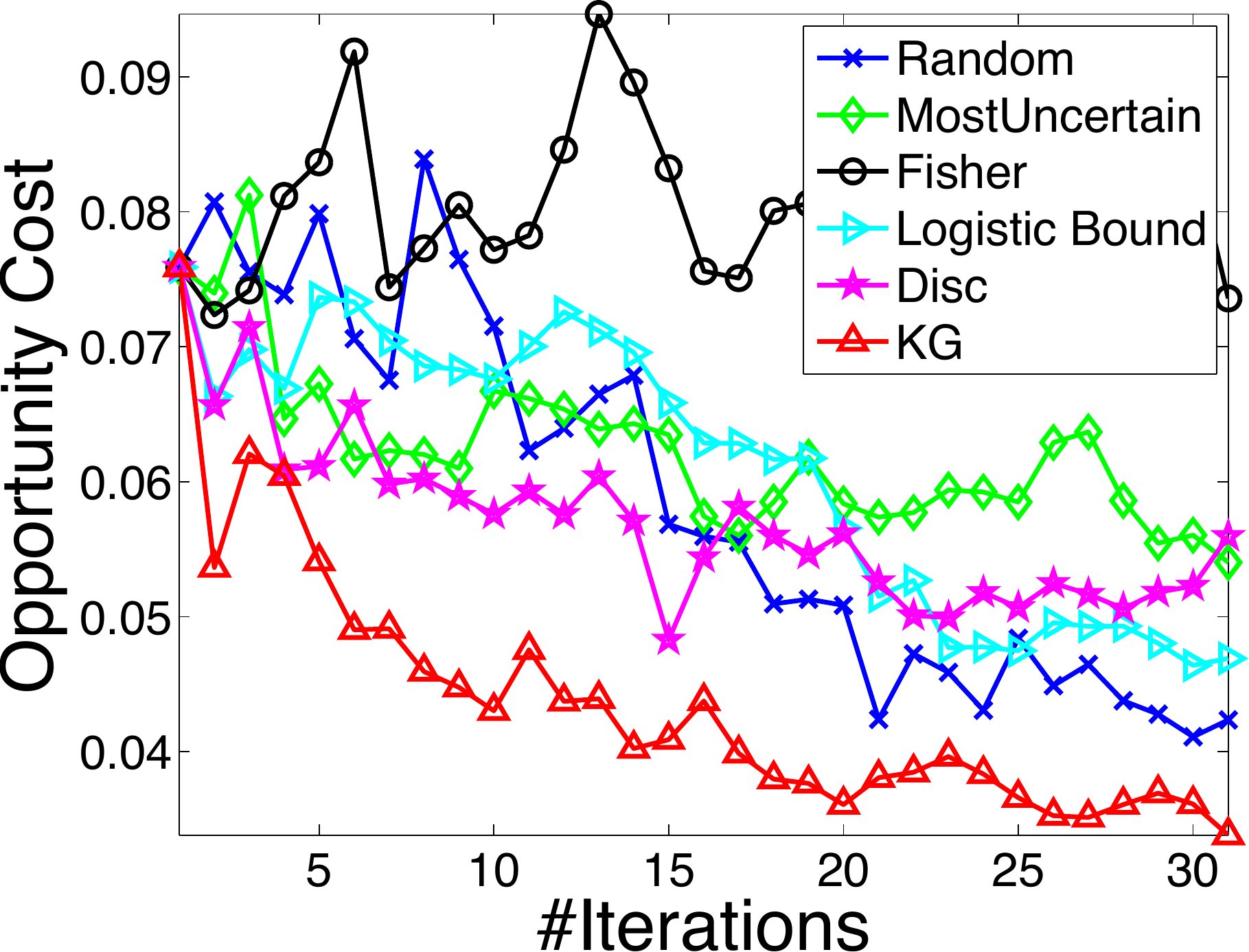} }

 \subfigure[glass identification]{
 \includegraphics[width=0.29\textwidth]{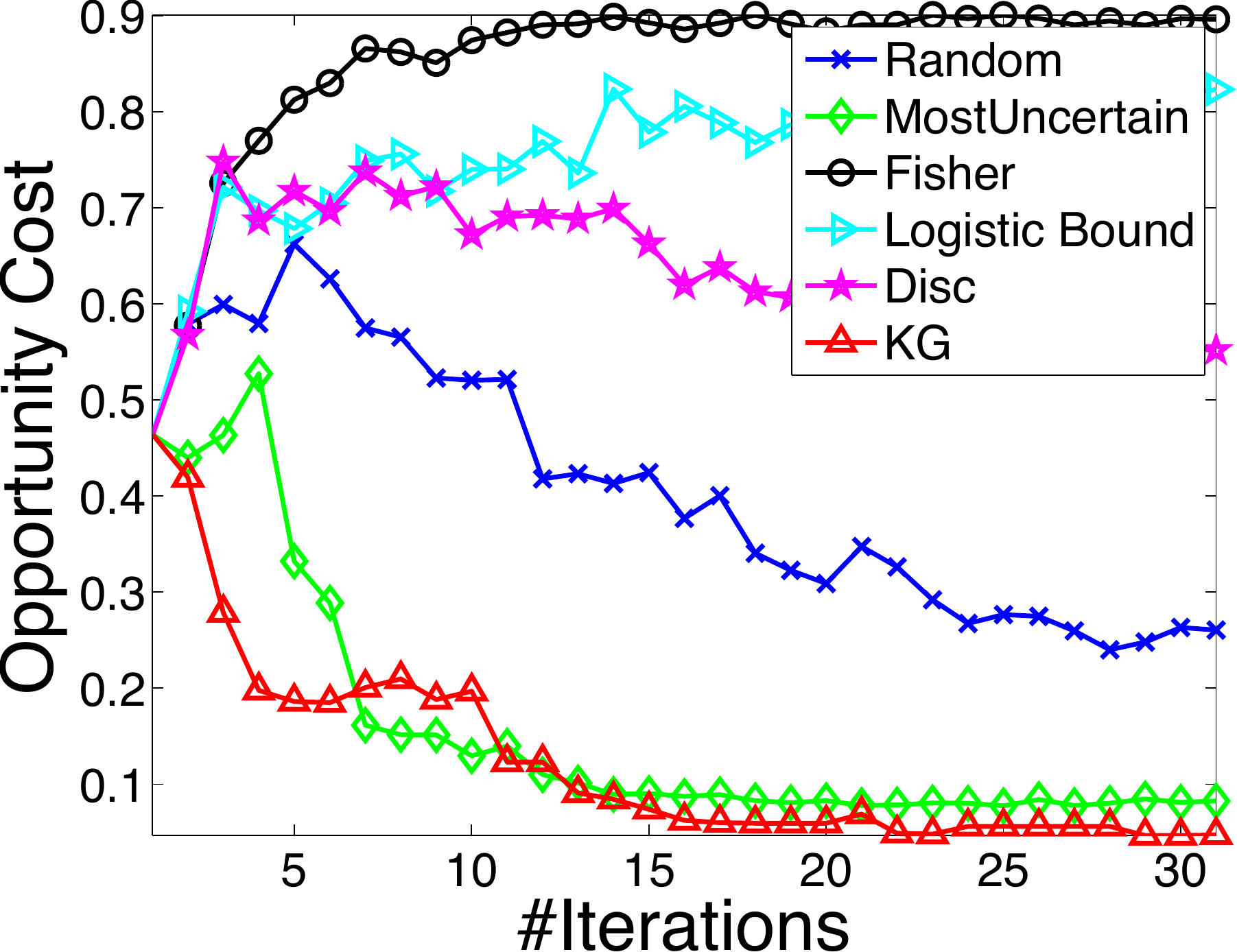} }

  \subfigure[blood transfusion]{
 \includegraphics[width=0.29\textwidth]{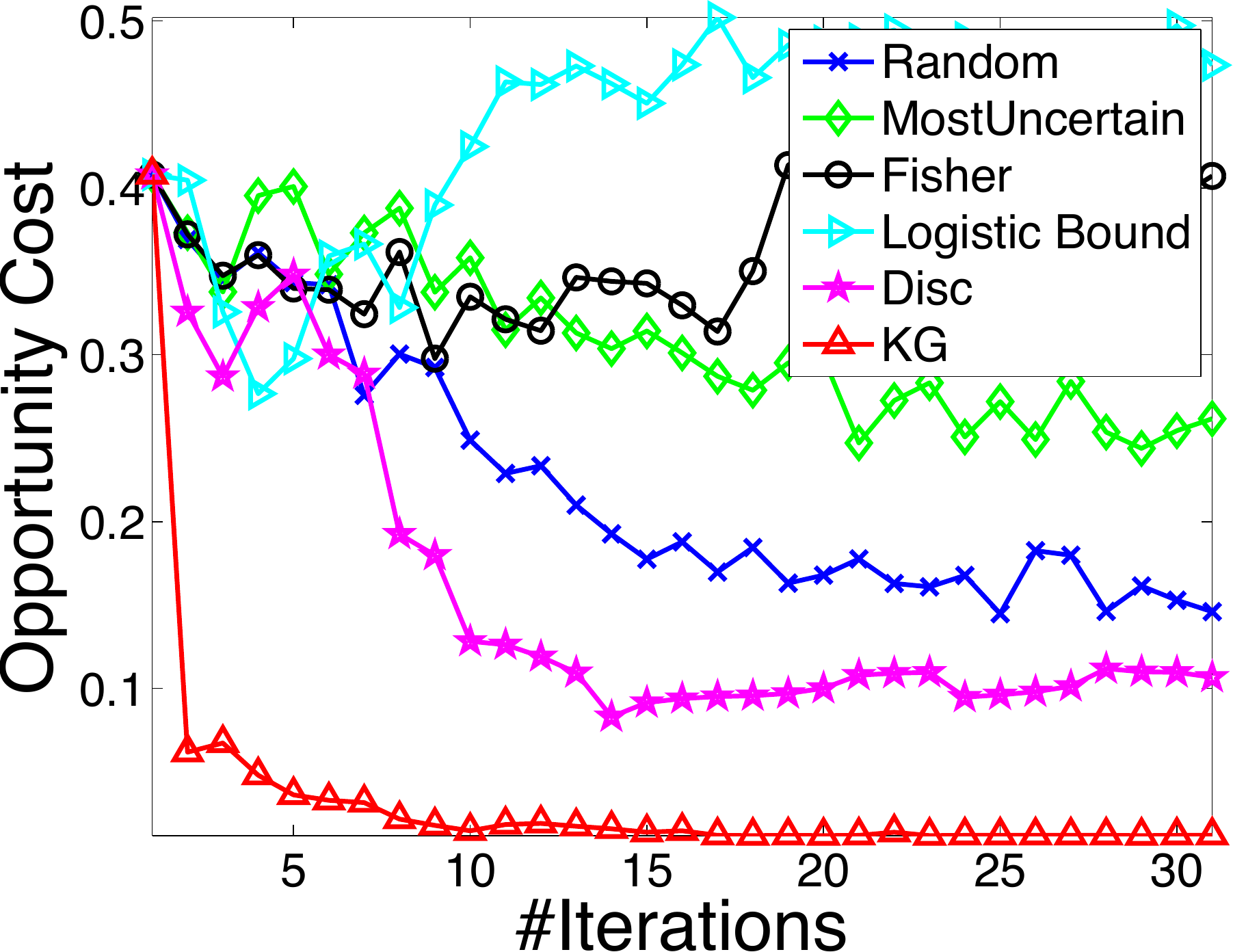} 
 } \\
 \subfigure[survival]{
 \includegraphics[width=0.29\textwidth]{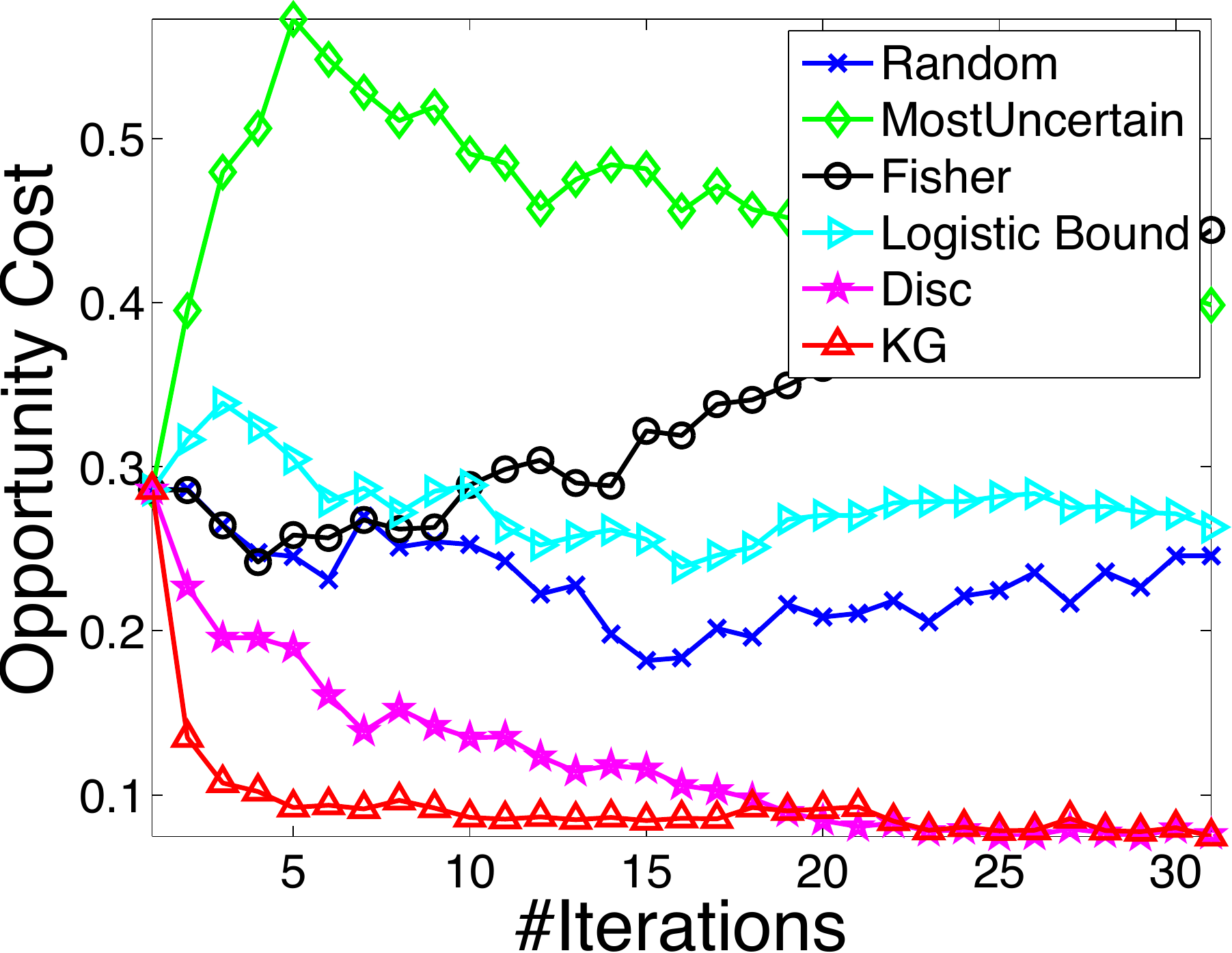} }

 \subfigure[breast cancer (wpbc)]{
 \includegraphics[width=0.29\textwidth]{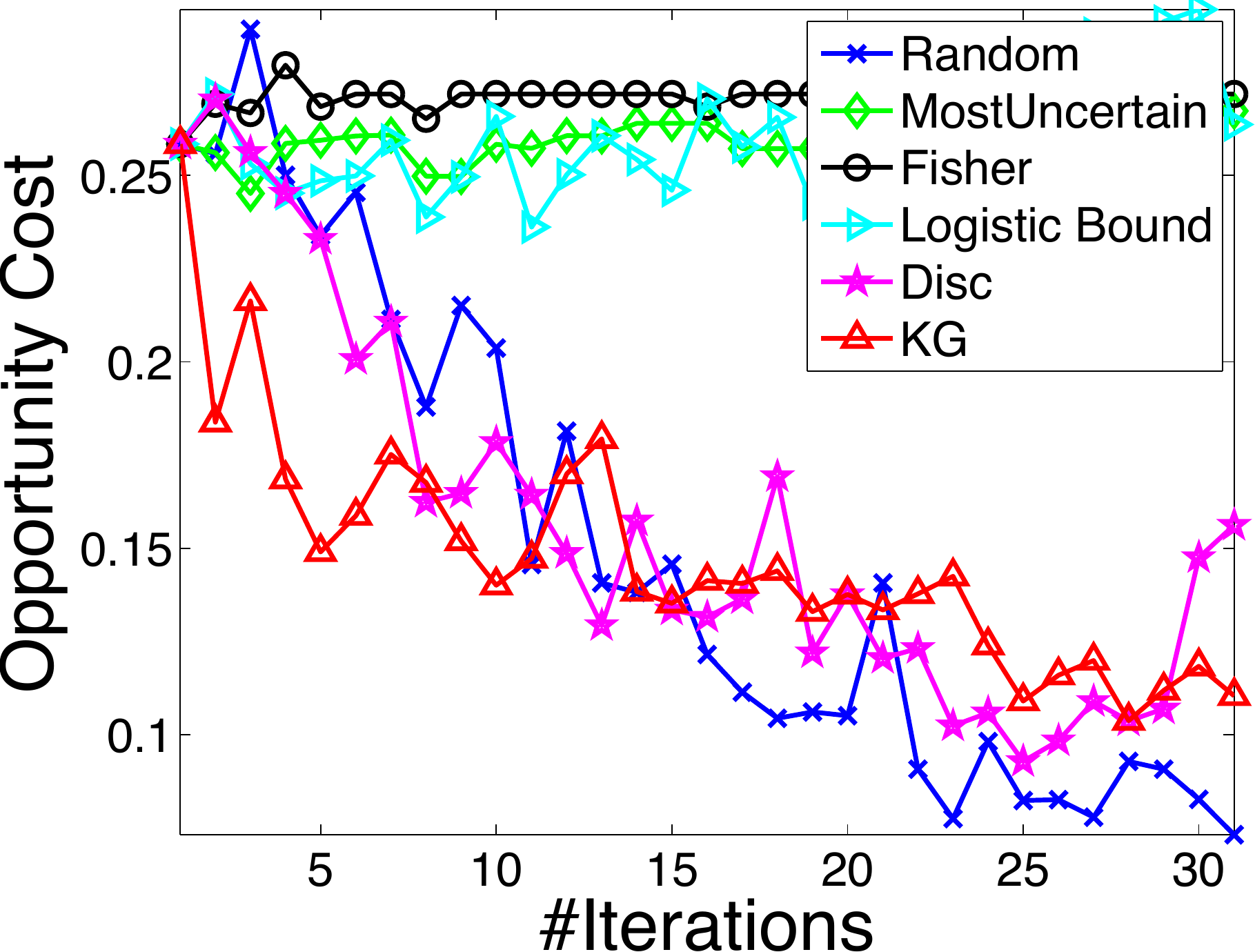} }

  \subfigure[planning relax ]{
 \includegraphics[width=0.29\textwidth]{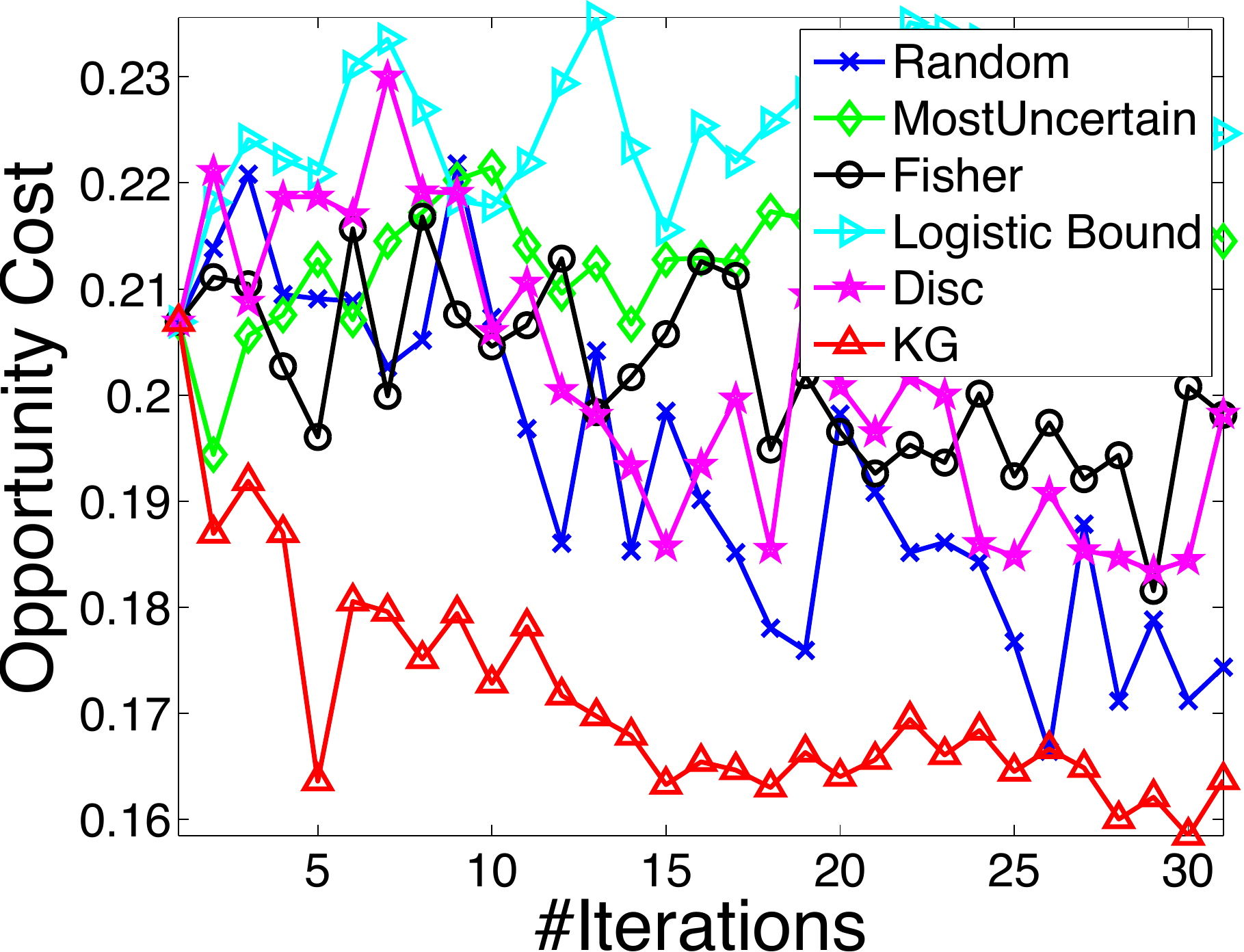} 
 } \\
  \subfigure[climate]{
 \includegraphics[width=0.29\textwidth]{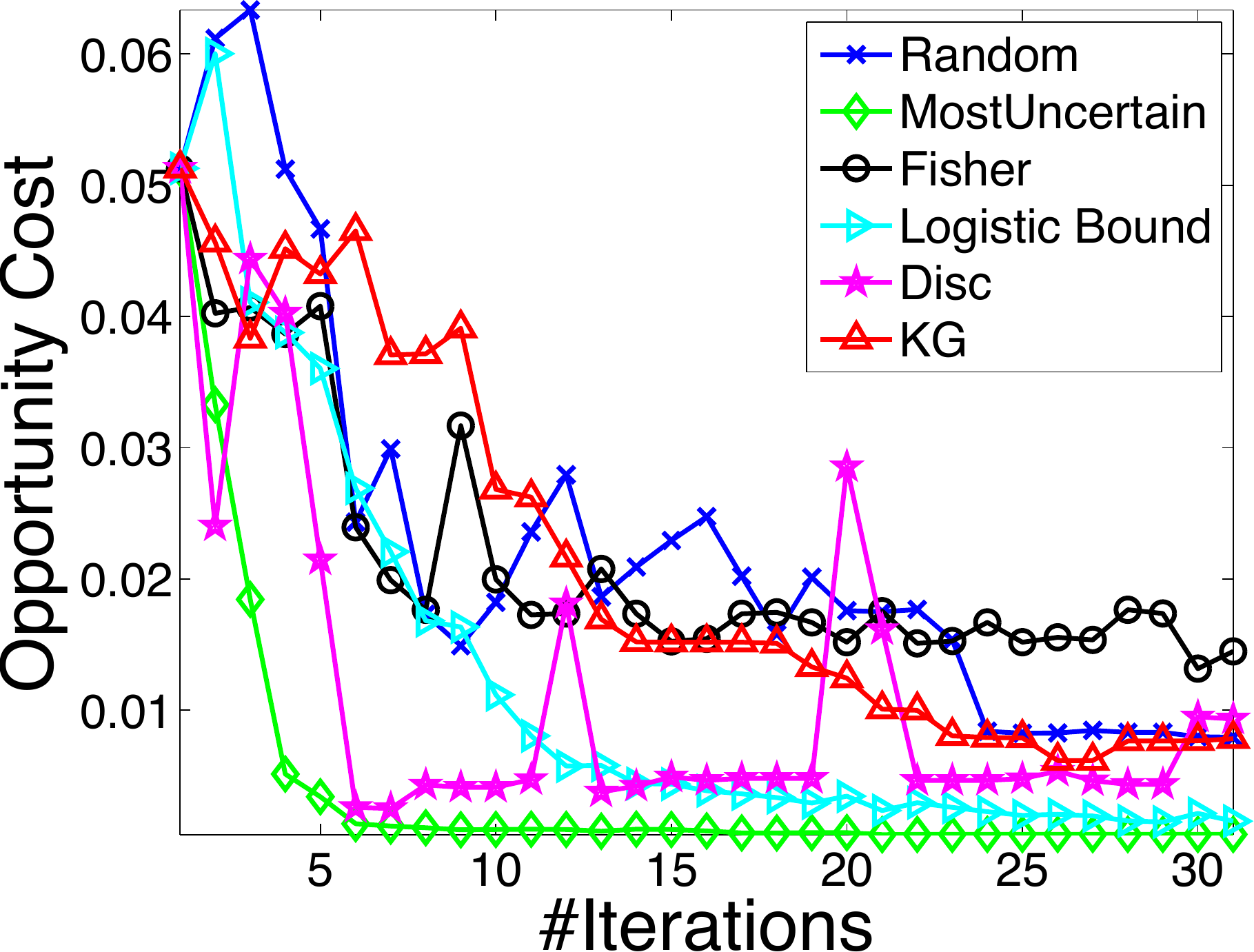} }

 \subfigure[Synthetic data, $d=10$]{
 \includegraphics[width=0.29\textwidth]{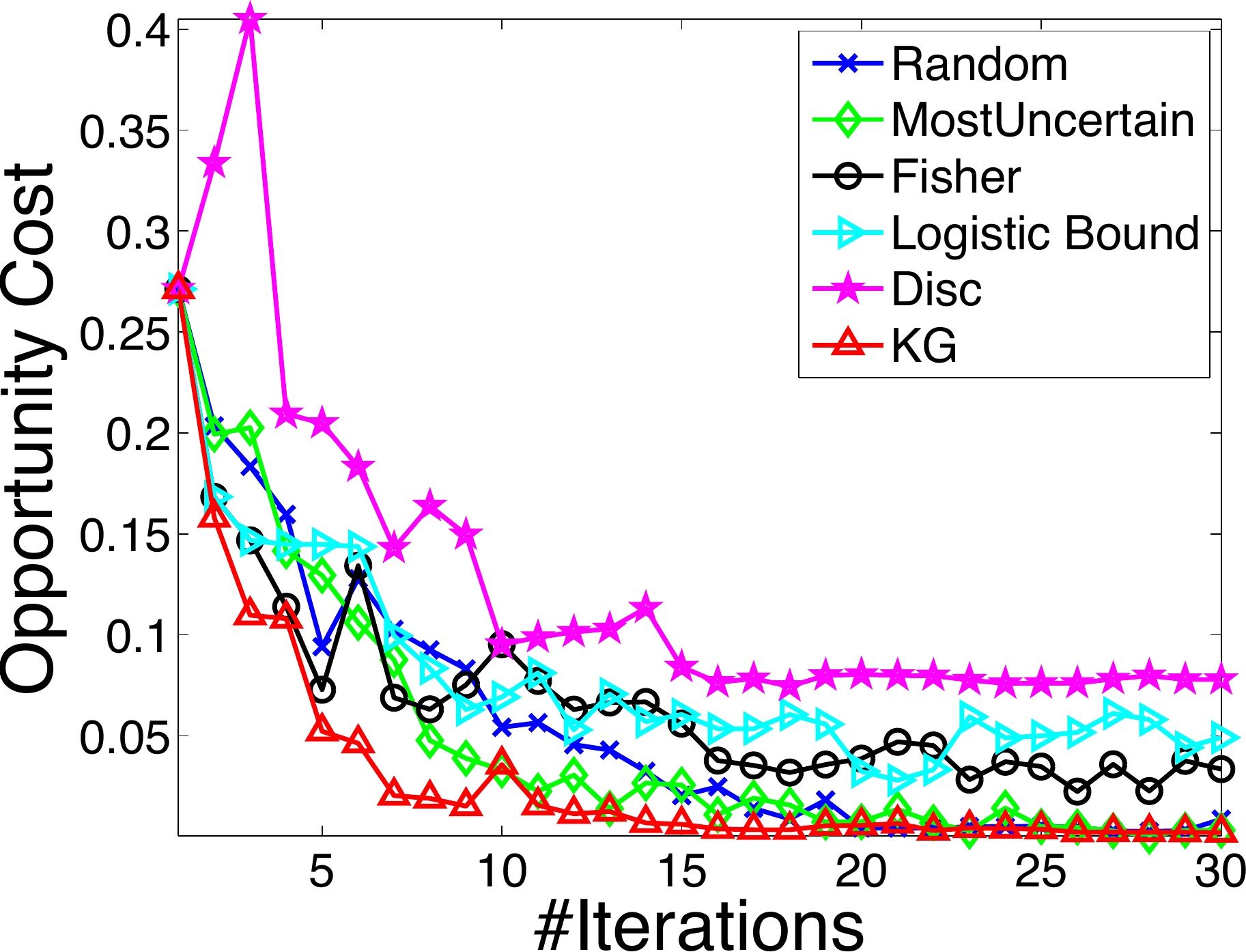} }

  \subfigure[Synthetic data, $d=15$]{
 \includegraphics[width=0.29\textwidth]{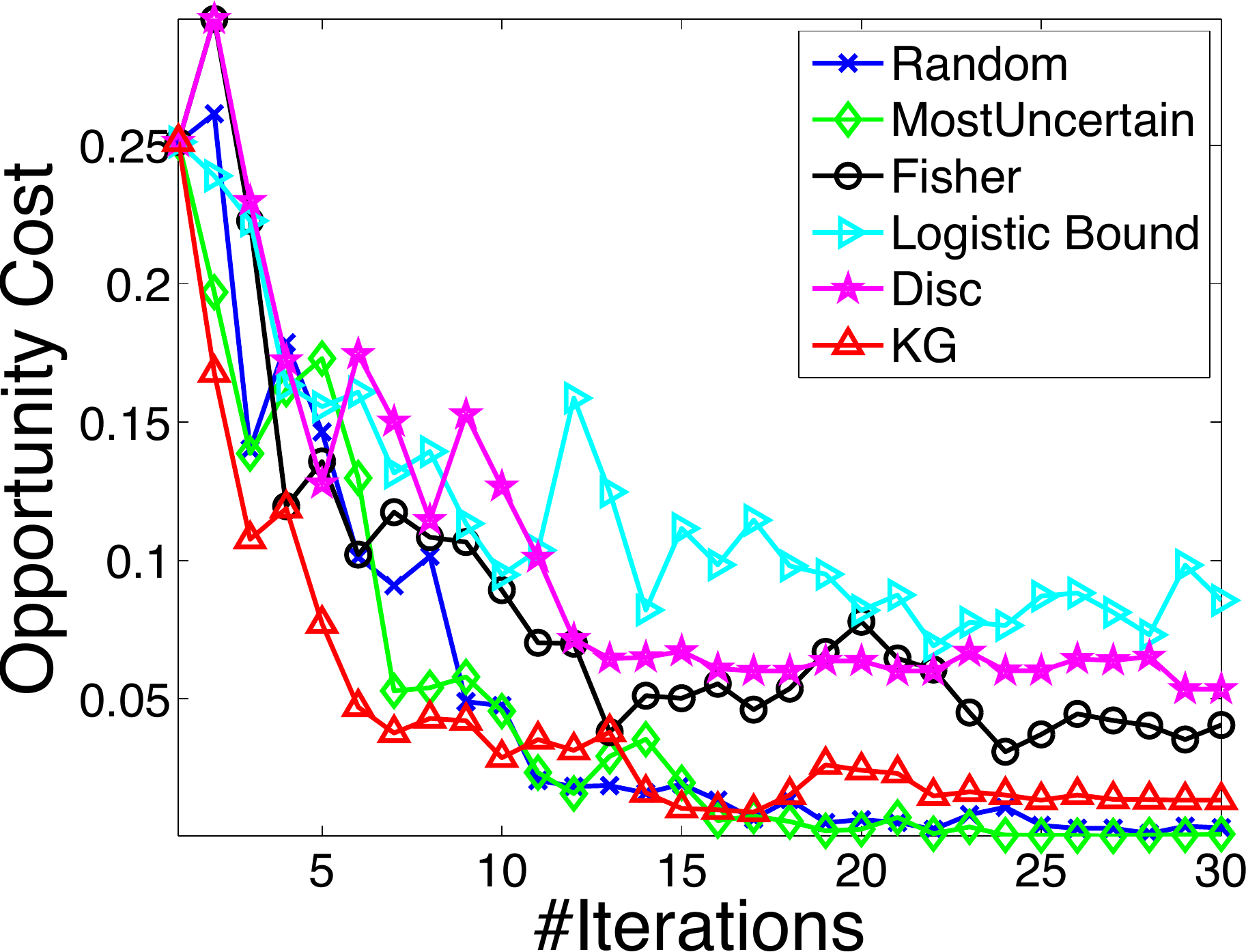} 
 } \\
     \end{tabular}
    \caption{Opportunity cost on UCI and synthetic datasets. \label{33}}
\end{figure}

Since there is no policy directly solving the same sequential response maximizing problem under a logistic model, considering the relationship with active learning as described in Section 1, we compare with the following state-of-the-art active learning policies compatible with logistic regression: 
Random sampling (Random), a myopic  method that selects the most uncertain instance each step (MostUncertain), Fisher information (Fisher) \cite{hoi2006batch}, the batch-mode active learning via error bound minimization (Logistic Bound) \cite{gu2014batch} and discriminative batch-mode active learning (Disc) \cite{guo2008discriminative} with batch size equal to 1. All the state transitions are based on recursive Bayesian logistic regression while different policies provides different rules for labeling decisions at each iteration. The experimental results are shown in figure \ref{33}. 
In all the figures, the x-axis denotes the number of measured alternatives and the y-axis represents the averaged opportunity cost  over 100 runs.

It is demonstrated in FIG. \ref{33} that KG outperforms the other policies significantly in most cases, especially in early iterations.  MostUncertain, Fisher and Logistic Bound perform well on  some datasets while badly on others.      Disc and Random yield relatively stable and satisfiable performance.  A possible explanation is that the goal of active leaning is to  learn a classifier which accurately predicts the labels of new examples so their criteria are not directly related  to maximize the response aside from the intent to learn the prediction.  After enough iterations when active learning methods presumably have the ability to achieve a good estimator of $\bm{w}^*$, their performance will be enhanced.
However, in the case when an experiment is expensive and only a small budget is allowed, the KG policy, which is designed specifically to maximize the response, is preferred.  


\section{Conclusion}
In this paper, we consider binary classification problems where we have to run expensive experiments, forcing us to learn the most from each experiment.  The goal is to learn the classification model as quickly as possible to identify the alternative with the highest response.  We develop a knowledge gradient policy using a logistic regression belief model, for which we developed an approximation method to overcome computational challenges in finding the knowledge gradient.  We provide a finite-time analysis on the estimated error, and report the results of a series of experiments that demonstrate its efficiency.

\bibliography{refer}

\end{document}